# Comparison of tree-based ensemble algorithms for merging satellite and earth-observed precipitation data at the daily time scale

Georgia Papacharalampous[1,*], Hristos Tyralis[2], Anastasios Doulamis[3], Nikolaos Doulamis[4]

[1] Department of Topography, School of Rural, Surveying and Geoinformatics Engineering, National Technical University of Athens, Iroon Polytechniou 5, 157 80 Zografou, Greece (papacharalampous.georgia@gmail.com, https://orcid.org/0000-0001-5446-954X)

[2] Department of Topography, School of Rural, Surveying and Geoinformatics Engineering, National Technical University of Athens, Iroon Polytechniou 5, 157 80 Zografou, Greece (montchrister@gmail.com, hristos@itia.ntua.gr, https://orcid.org/0000-0002-8932-4997)

[3] Department of Topography, School of Rural, Surveying and Geoinformatics Engineering, National Technical University of Athens, Iroon Polytechniou 5, 157 80 Zografou, Greece (adoulam@cs.ntua.gr, https://orcid.org/0000-0002-0612-5889)

[4] Department of Topography, School of Rural, Surveying and Geoinformatics Engineering, National Technical University of Athens, Iroon Polytechniou 5, 157 80 Zografou, Greece (ndoulam@cs.ntua.gr, https://orcid.org/0000-0002-4064-8990)

* Corresponding author

**Abstract**: Merging satellite products and ground-based measurements is often required for obtaining precipitation datasets that simultaneously cover large regions with high density and are more accurate than pure satellite precipitation products. Machine and statistical learning regression algorithms are regularly utilized in this endeavour. At the same time, tree-based ensemble algorithms are adopted in various fields for solving regression problems with high accuracy and low computational cost. Still, information on which tree-based ensemble algorithm to select for correcting satellite precipitation products for the contiguous United States (US) at the daily time scale is missing from the literature. In this study, we worked towards filling this methodological gap by conducting an extensive comparison between three algorithms of the category of interest, specifically between random forests, gradient boosting machines (gbm) and extreme gradient boosting (XGBoost). We used daily data from the PERSIANN (Precipitation Estimation from Remotely Sensed Information using Artificial Neural Networks) and the IMERG (Integrated Multi-satellitE Retrievals for GPM) gridded datasets. We also used earth-observed precipitation data from the Global Historical Climatology Network daily (GHCNd) database. The experiments referred to the entire contiguous US and additionally included the application of the linear regression algorithm for benchmarking purposes. The results suggest that XGBoost is the best-performing tree-based ensemble algorithm among those compared. Indeed, the mean relative improvements that it provided with respect to linear regression (for the case that the latter algorithm was run with the same predictors as XGBoost) are equal to 52.66%, 56.26% and 64.55% (for three different predictor sets), while the respective values are 37.57%, 53.99% and 54.39% for random forests, and 34.72%, 47.99% and 62.61% for gbm. Lastly, the results suggest that IMERG is more useful than PERSIANN in the context investigated.

**Keywords**: contiguous US; gradient boosting machines; IMERG; machine learning; PERSIANN; random forests; remote sensing; satellite precipitation correction; spatial interpolation; XGBoost

## 1. Introduction

Machine and statistical learning algorithms (e.g., those documented in Hastie et al. 2009; James et al. 2013; Efron and Hastie 2016) are increasingly adopted for solving a variety of practical problems in hydrology (Dogulu et al. 2015; Xu et al. 2018; Quilty et al. 2019; Curceac et al. 2020; Quilty and Adamowski 2020; Rahman et al. 2020; Althoff et al. 2021; Fischer and Schumann 2021; Cahyono 2022; Mehedi et al. 2022; Rozos et al. 2022a,b; Papacharalampous and Tyralis 2022b; Granata et al. 2023; Payne et al. 2023) and beyond (Goetz et al. 2015; Bahl et al. 2018; Feng et al. 2020; Rustam et al. 2021; Bamisile et al. 2022). Among the entire pool of such algorithms, the tree-based ensemble ones (i.e., those combining decision trees under properly designed ensemble learning strategies; Sagi and Rokach 2018) are of special interest for many practical problems, as most of their software implementations offer high predictive performance with low computational cost, together with high automation and some degree of explainability (Tyralis et al. 2019; Tyralis and Papacharalampous 2021). Additionally, they usually do not require extensive preprocessing and hyperparameter tuning to perform well (Hastie et al. 2009; Tyralis and Papacharalampous 2021). On the other hand, as they are highly flexible algorithms, they are less interpretable than simpler algorithms (e.g., linear regression), due to the well-recognized trade-off between interpretability and flexibility (James et al. 2013).

Notably, the known theoretical properties of the various tree-based ensemble algorithms (including random forests, gradient boosting machines – gbm and extreme gradient boosting – XGBoost; Breiman 2001, Friedman 2001, Chen and Guestrin 2016) cannot support the selection of the most appropriate one among them for each practical problem. Instead, such a selection could rely on attentively designed empirical comparisons. Thus, such comparisons of tree-based ensemble algorithms are conducted with increasing frequency in various scientific fields (Fan et al. 2018; Besler et al. 2019; Ahmad and Zhang 2020; Liu et al. 2020; Ziane et al. 2021; Park and Kim 2022).

Tree-based ensemble algorithms are regularly applied and compared to other machine and statistical learning algorithms for the task of merging satellite products and ground-based measurements. This task is the general focus of this work, together with the general concept of tree-based ensemble algorithms, and is commonly executed in the literature in the direction of obtaining precipitation datasets that cover large geographical regions with high density and, simultaneously, are more accurate than pure satellite precipitation products. The importance of this same task can be perceived through the inspection of the major research topics appearing in the hydrological literature (see, e.g., those discussed in Montanari et al. 2013, Blöschl et al. 2019). Relevant examples of applications and comparisons are available in He et al. (2016), Baez-Villanueva et al. (2020), Chen et al. (2021), Zhang et al. (2021), Fernandez-Palomino et al. (2022), Lei et al. (2022) and Militino et al. (2023).

These examples refer to various temporal resolutions and many different geographical regions around the globe (see also the reviews by Hu et al. 2019 and Abdollahipour et al. 2022), with the daily temporal resolution and the Unites States (US) being frequent cases. Nonetheless, a relevant comparison of tree-based ensemble algorithms for the latter temporal resolution and the latter geographical region is missing from the literature, with the closest investigations at the moment being those available in the work by Lei et al.

(2022), which focuses on China. In this work, we fill this specific literature gap. Notably, the selection of the most accurate regression algorithm from the tree-based ensemble family could be particularly useful at the daily temporal scale, in which the size of the datasets for large geographical areas might impose significant limitations on the application of other accurate machine and statistical learning regression algorithms due to their large computational costs.

## 2. Methods

Random forests, gbm and XGBoost were applied in a cross-validation setting (see Section 3.2) for conducting an extensive comparison in the context of merging gridded satellite products and ground-based measurements at the daily time scale. Additionally, the linear regression algorithm was applied in the same setting for benchmarking purposes. In this section, we provide brief descriptions of the four afore-mentioned algorithms. Extended descriptions are out of the scope of this work, as they are widely available in the machine and statistical learning literature (e.g., in Hastie et al. 2009; James et al. 2013; Efron and Hastie 2016). Statistical software information that ensures the work's reproducibility is provided in Appendix A.

### 2.1 Linear regression

The results of this work are reported in terms of relative scores (see Section 3.3). These scores were computed with respect to the linear regression algorithm, which models the dependent variable as a linear weighted sum of the predictor variables (Hastie et al. 2009, pp 43–55). A squared error scoring function facilitated this algorithm's fitting.

### 2.2 Random forests

Random forests (Breiman 2001) are the most commonly used algorithm in the context of merging gridded satellite products and ground-based measurements (see the examples in Hengl et al. 2018). A detailed description of this algorithm can be found in Tyralis et al. (2019b), along with a systematic review of its application in water resources. Notably, random forests are an ensemble learning algorithm and, more precisely, an ensemble of regression trees that is based on bagging (acronym for "bootstrap aggregation") but with an additional randomization procedure. The latter aims at reducing overfitting. In this work, random forests were implemented with all their hyperparameters kept as default. For instance, the number of trees was equal to 500. This methodological choice is adequate, as random forests are known to perform well without tuning as long as they are applied with a large number of trees (Tyralis et al. 2019b).

## 2.3 Gradient boosting machines

Another ensemble learning algorithm that was herein used with regression trees as base learners is gbm (Friedman 2001, Mayr et al. 2014). The main concept behind this ensemble algorithm and, more generally, behind all the boosting algorithms (including the one described in Section 2.4) is the iterative training of new base learners using the errors of previously trained base learners (Natekin and Knoll 2013, Tyralis and Papacharalampous 2021). For gradient boosting machines, a gradient descent algorithm adapts the loss function for achieving optimal fitting. This loss function is the squared error scoring function herein. Consistency in respect to the implementation of random forests was ensured by setting the number of trees equal to 500. The remaining hyperparameters were kept as default. This latter methodological choice is expected to be adequate, as boosting procedures are designed with the ability to run as "off the shelf" procedures (Tyralis and Papacharalampous 2021).

## 2.4 Extreme gradient boosting

XGBoost (Chen and Guestrin 2016) is the third tree-based ensemble learning and the second boosting algorithm implemented in this work. In the implementations of this work, all the hyperparameters were kept as default (as this is expected to be adequate; see the above section), except for the maximum number of iterations that were set to 500.

Aside from applying XGBoost in a cross-validation setting for its comparison to the remaining algorithms, we also utilized it with the same hyperparameter values for ensuring some degree of explainability in terms of variable importance under the more general explainable machine learning culture (Linardatos et al. 2020, Roscher et al. 2020, Belle and Papantonis 2021). Specifically, we computed the gain importance metric, which is available in the XGBoost algorithm. This metric assesses the *"fractional contribution of each feature to the model based on the total gain of this feature's splits"*, with higher values indicating more important features (Chen et al. 2022c).

## 3. Data and application

### 3.1 Data

For our experiments, we retrieved and used daily earth-observed precipitation, gridded satellite precipitation and elevation data for the gauged locations and grid points shown in Figures 1 and 2.

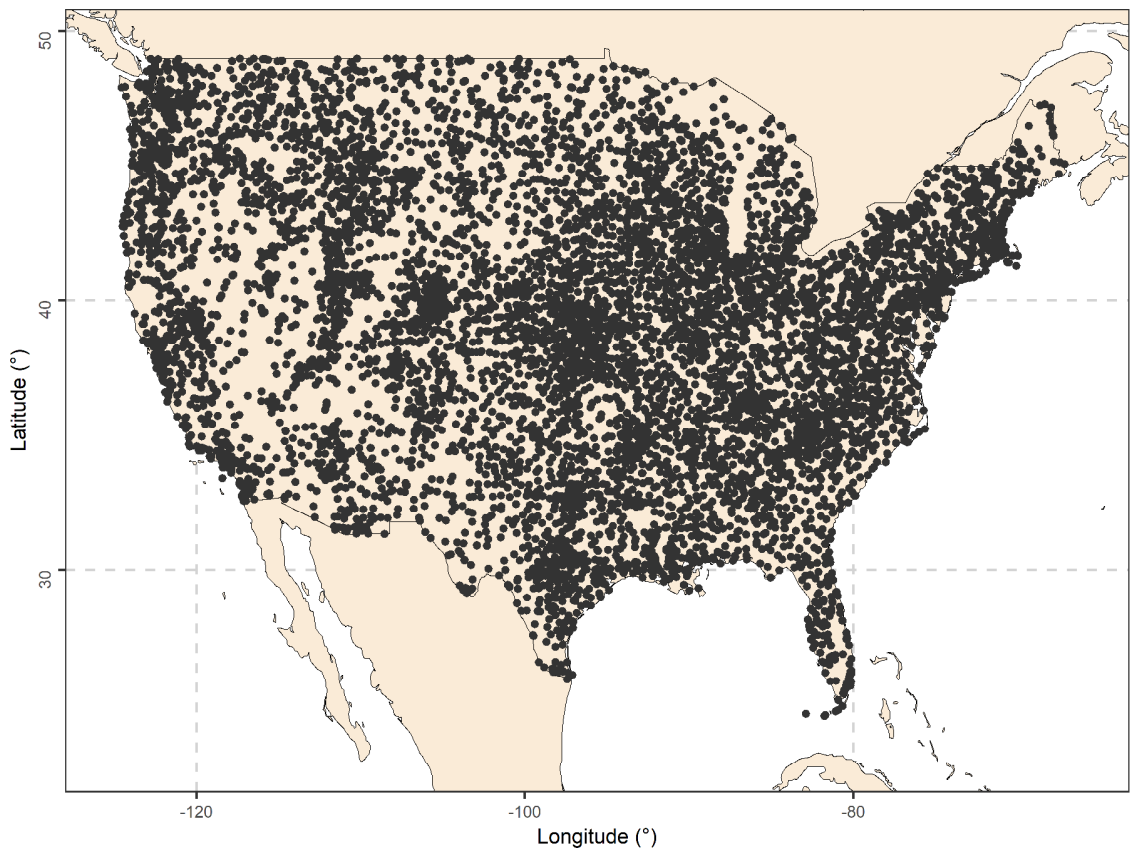

Figure 1. Map of the geographical locations of the earth-located stations that offered data for this work.

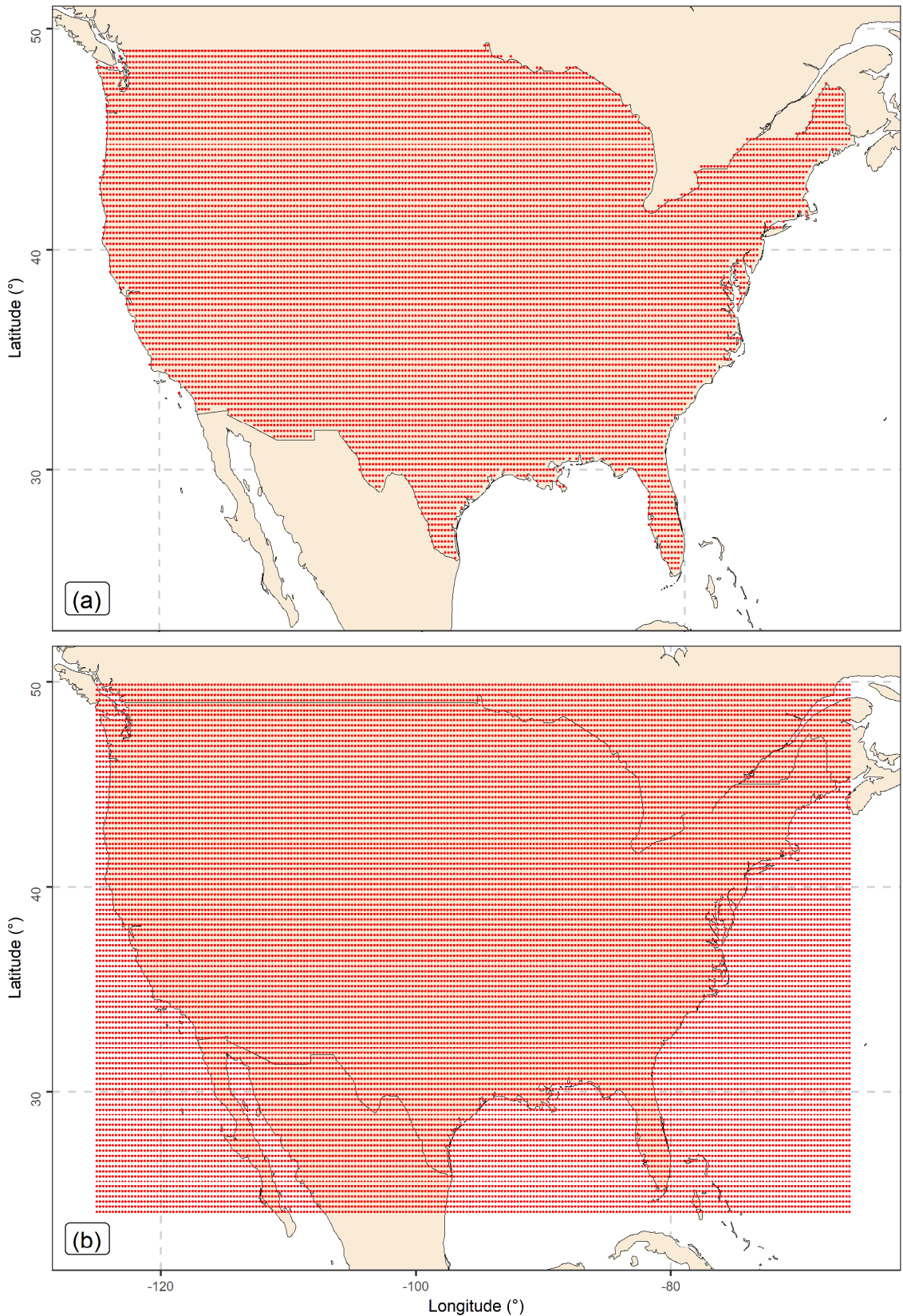

Figure 2. Maps of the geographical locations of the points composing the (a) PERSIANN and (b) IMERG grids utilized in this work.

### 3.1.1 Earth-observed precipitation data

Daily precipitation totals from the Global Historical Climatology Network daily (GHCNd) (Durre et al. 2008, 2010, Menne et al. 2012) were used for comparing the algorithms. More precisely, data from 7 264 earth-located stations spanning across the contiguous US (see Figure 1) were extracted. These data cover the two-year time period 2014–2015. Data retrieval was made from the website of the NOAA National Climatic Data Center (https://www1.ncdc.noaa.gov/pub/data/ghcn/daily; accessed on 2022-02-27).

### 3.1.2 Satellite precipitation data

For comparing the algorithms, we additionally used gridded satellite daily precipitation data from the current operational PERSIANN (Precipitation Estimation from Remotely Sensed Information using Artificial Neural Networks) system (see the geographical locations of the extracted PERSIANN grid with a spatial resolution of 0.25 degree x 0.25 degree in Figure 2a) and the GPM IMERG (Integrated Multi-satellitE Retrievals) Late Precipitation L3 1 day 0.1 degree x 0.1 degree V06. These two gridded satellite precipitation databases were developed by the Centre for Hydrometeorology and Remote Sensing (CHRS) at the University of California, Irvine (UCI) and the National Aeronautics and Space Administration (NASA) Goddard Earth Sciences (GES) Data and Information Services Center (DISC), respectively. More precisely, the PERSIANN data were retrieved from the website of CHRS (https://chrsdata.eng.uci.edu; accessed on 2022-03-07) and the IMERG data were retrieved from the website of NASA Earth Data (https://doi.org/10.5067/GPM/IMERGDL/DAY/06; accessed on 2022-12-10). The extracted data cover the entire contiguous US for the two-year time period 2014–2015. Notably, the extracted PERSIANN data were used in the experiments with their original spatial resolution, while the extracted GPM IMERG data were used for forming data with a spatial resolution of 0.25 degree x 0.25 degree by applying bilinear interpolation on the CMORPH0.25 grid. The herein formed data and their grid (see Figure 2b) are those referred to in what follows as "IMERG values" and "IMERG grid", respectively, and are the ones used in the experiments.

### 3.1.3 Elevation data

Elevation is a key predictor variable for many hydrological processes (Xiong et al. 2022). Therefore, we estimated its value for all the geographical locations shown in Figure 1. For this estimation, we relied on the Amazon Web Services (AWS) Terrain Tiles

([https://registry.opendata.aws/terrain-tiles](https://registry.opendata.aws/terrain-tiles); accessed on 2022-09-25).

## 3.2 Validation setting and predictor variables

To formulate the regression settings of this work, we first defined earth-observed daily total precipitation at a point of interest (which could be station 1 in Figure 3) as the dependent variable. Then, we adopted procedures proposed in Papacharalampous et al. (2023) to compute the observations of possible predictor variables. Separately for each of the two satellite precipitation grids (see Figure 2), we determined the closest four grid points to each ground-based station from those depicted in Figure 1. We also computed the distances $d_i$, $i$ = 1, 2, 3, 4 from these grid points and indexed the latter as $S_i$, $i$ = 1, 2, 3, 4 based on the following order: $d_1 < d_2 < d_3 < d_4$ (see Figure 3). Throughout this work, the distances $d_i$, $i$ = 1, 2, 3, 4 are also respectively called "PERSIANN distances 1–4" or "IMERG distances 1–4" (depending on whether we refer to the PERSIANN grid or the IMERG grid) and the daily precipitation values at the grid points 1–4 are called "PERSIANN values 1–4" or "IMERG values 1–4" (depending on whether we refer to the PERSIANN grid or the IMERG grid).

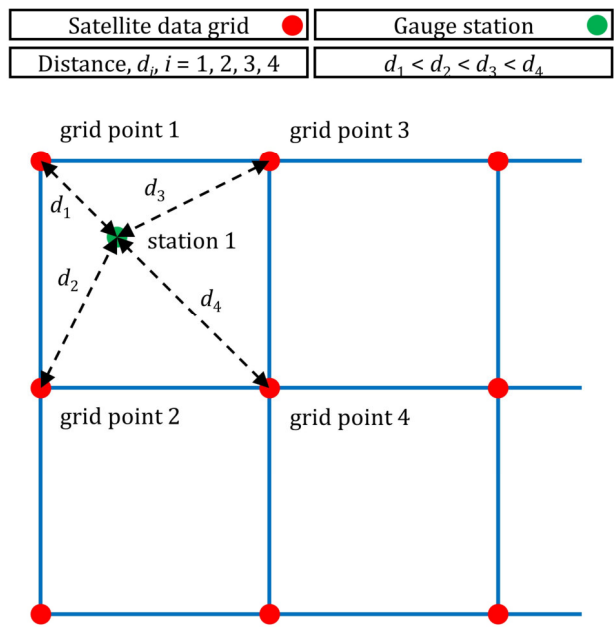

Figure 3. Setting of the regression problem. Note that the term "grid point" is used to describe the geographical locations with satellite data, while the term "station" is used to describe the geographical locations with ground-based measurements. Note also that, throughout this work, the distances $d_i$, $i$ = 1, 2, 3, 4 are also respectively called "PERSIANN distances 1–4" or "IMERG distances 1–4" (depending on whether we refer to the PERSIANN grid or the IMERG grid) and the daily precipitation values at the grid points 1–4 are called "PERSIANN values 1–4" or "IMERG values 1–4" (depending on whether we refer to the PERSIANN grid or the IMERG grid).

Based on the above, the predictor variables for the technical problem of interest could include the PERSIANN values 1–4, the IMERG values 1–4, the PERSIANN distances 1–4, the IMERG distances 1–4 and the station's elevation. We defined and examined three sets of predictor variables (see Table 1). Each of them defines a different regression setting that includes 4 833 007 samples. These samples were exploited under a two-fold cross-validation scheme for comparing the three tree-based ensemble algorithms outlined in Section 2 in the context of merging gridded satellite precipitation products and ground-based precipitation measurements at the daily temporal scale. The same samples were explored by estimating the Spearman correlation (Spearman 1904) for the various pairs of variables and by ranking the predictor variables based on their importance in the regression. The latter methodological step was made by applying explainable machine learning procedures offered by the XGBoost algorithm (see Section 2.4).

Table 1. Inclusion of predictor variables in the predictor sets examined in this work.

| Predictor variable | Predictor set 1 | Predictor set 2 | Predictor set 3 |
|---|---|---|---|
| PERSIANN value 1 | ✓ | × | ✓ |
| PERSIANN value 2 | ✓ | × | ✓ |
| PERSIANN value 3 | ✓ | × | ✓ |
| PERSIANN value 4 | ✓ | × | ✓ |
| IMERG value 1 | × | ✓ | ✓ |
| IMERG value 2 | × | ✓ | ✓ |
| IMERG value 3 | × | ✓ | ✓ |
| IMERG value 4 | × | ✓ | ✓ |
| PERSIANN distance 1 | ✓ | × | ✓ |
| PERSIANN distance 2 | ✓ | × | ✓ |
| PERSIANN distance 3 | ✓ | × | ✓ |
| PERSIANN distance 4 | ✓ | × | ✓ |
| IMERG distance 1 | × | ✓ | ✓ |
| IMERG distance 2 | × | ✓ | ✓ |
| IMERG distance 3 | × | ✓ | ✓ |
| IMERG distance 4 | × | ✓ | ✓ |
| Station elevation | ✓ | ✓ | ✓ |

### 3.3 Performance metrics and assessment

The performance assessment relied on procedures proposed by Papacharalampous et al. (2023). These procedures are reported in what follows. First, we computed the median of the squared error function, separately for each set {algorithm, predictor set, test fold}. Note that the squared error scoring function can adequately support our performance comparisons, as it is consistent for the mean functional of the predictive distributions (Gneiting 2011). Subsequently, two relative scores (which are else referred to as "relative improvements" throughout this work) were computed for each set {algorithm, predictor

set}. For that, the two median squared error (MedSE) values offered by each set {algorithm, predictor set} (each corresponding to a different test fold) were utilized, together with their corresponding MedSE values offered by the reference modelling approach, which was defined as the linear regression when run with the same predictor set as the modelling approach to which the relative score referred. More precisely, the relative score was computed as the difference between the score of the set {algorithm, predictor set} minus the score of the reference modelling approach, multiplied by 100 and divided by the score of the reference modelling approach. Then, mean relative scores (which are also referred to as "mean relative improvements" throughout this work) were computed by averaging, separately for each set {algorithm, predictor set}, the relative scores. The procedures for computing the relative scores and the mean relative scores were repeated by considering the set {linear regression, predictor set 1} as the reference modelling approach for all the sets {algorithm, predictor set}.

Mean rankings of the machine and statistical learning algorithms were also computed. For that, and separately for each set {case, predictor set} (with each case belonging to one test fold only), we first ranked the four algorithms based on their squared errors. Then, we averaged these rankings, separately for each set {predictor set, test fold}. Lastly, we obtained the mean rankings reported by averaging the two previously computed mean ranking values corresponding to the same predictor set. We also computed the rankings collectively for all the predictor sets.

## 4. Results

### 4.1 Regression setting exploration

Regression setting explorations can facilitate interpretations of the results of prediction experiments, at least to some extent. Therefore, in Figure 4, we present the Spearman correlation estimates for the various variable pairs appearing in the regression settings examined in this work. As it could be expected, the magnitude of the relationships between the predictand (i.e., the precipitation quantity observed at the earth-located stations, which is referred to as "true value" in Figure 4) and the 17 predictor variables seems to depend, to some extent, on the satellite rainfall product. Indeed, the Spearman correlation estimates made for the relationships between the predictand and precipitation quantities from the IMERG grid are equal to 0.45, while the corresponding estimates for the case of the RERSIANN grid are equal to 0.40. The remaining relationships between the predictand and predictor variables are far less intense, almost negligible, based on the Spearman correlation statistic. Still, they could also provide information in the regression settings.

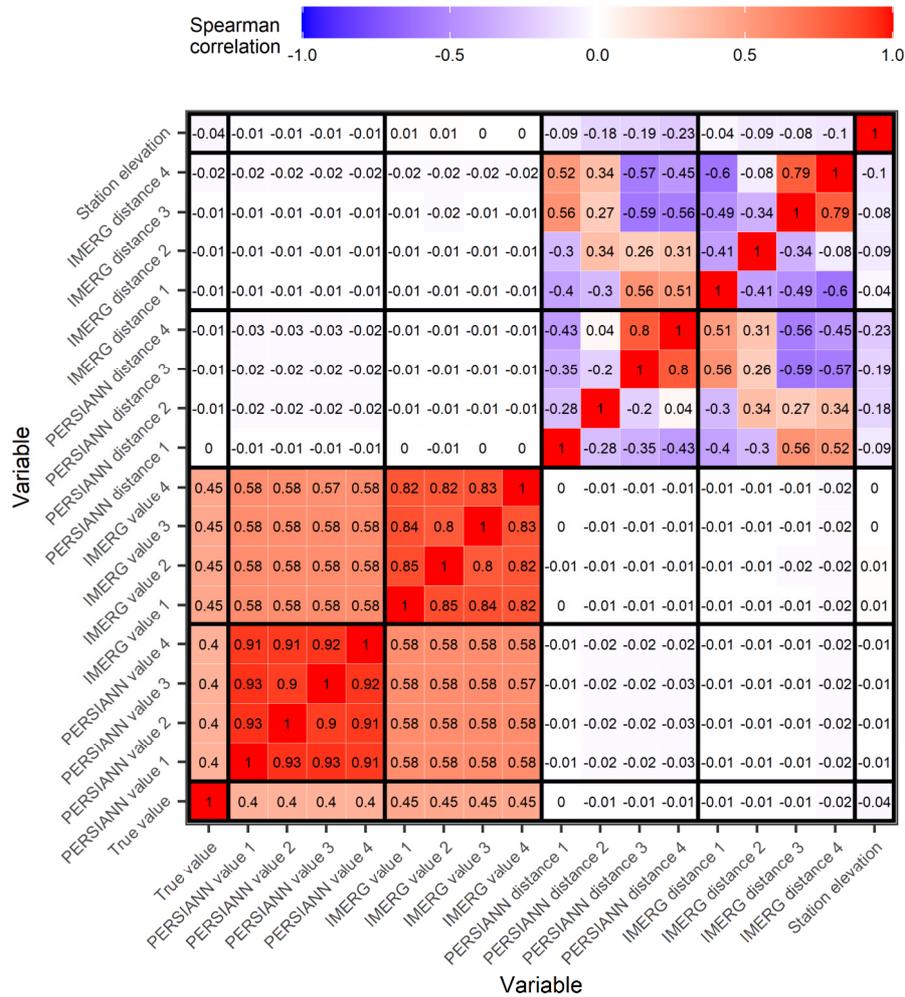

Figure 4. Heatmap of the Spearman correlation estimates for the various variable pairs appearing in the regression settings of this work.

The relationships between the predictor variables also exhibit various intensities. The most intense among them, according to the Spearman correlation statistic, are the relationships between the PERSIANN values, for which the estimates obtained are equal to 0.90, 0.91, 0.92 and 0.93. The relationships between the IMERG values are also intense, with the corresponding Spearman correlation estimates being equal to 0.80, 0.82, 0.83, 0.84 and 0.85. The Spearman correlation estimates referring to the relationships between the distances, as well as those referring to the relationships between the distances and the earth-located station's elevation (with the latter being referred to as "station elevation" in the visualizations), are either positive or negative and smaller (in absolute terms) than the Spearman correlation estimates referring to the relationships between the PERSIANN values and the relationships between the IMERG values. Still, some of them are of similar magnitude as those referring to the relationships between the PERSIANN and IMERG values.

Furthermore, Figure 5 presents the importance scores and rankings computed for the 17 predictor variables through the XGBoost algorithm and by considering all of these predictor variables in the regression setting. The four IMERG values were found to be the most important predictor variables. Moreover, the fifth and sixth most important predictors are PERSIANN value 1 and station elevation, respectively, and PERSIANN values 2–4 follow in the line, while the eight distances are the eight least important predictor variables. Notably, the fact that station elevation is more important than three of the four PERSIANN values could not be expected by inspecting the Spearman correlation estimates (see again Figure 4).

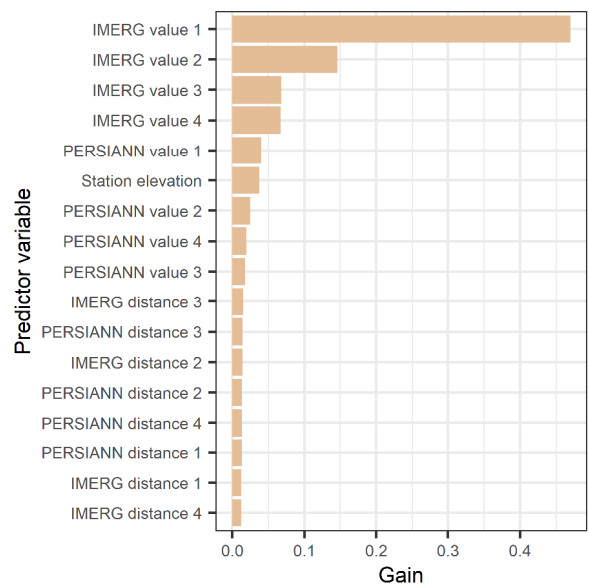

Figure 5. Barplot of the gain scores computed for the predictor variables by utilizing the extreme gradient boosting algorithm. The predictor variables are presented from the most to the least important ones (from top to bottom) based on the same scores.

## 4.2 Algorithm comparison

Figure 6 facilitates the comparison of the four machine and statistical learning algorithms in terms of the square error function, separately for each predictor set. The mean relative improvements (see Figure 6a) suggest that XGBoost is the best algorithm for all the predictor sets. For predictor set 1 (which incorporates, among others, information from the PERSIANN gridded precipitation dataset; see Table 1), random forests exhibit very similar performance compared to that of XGBoost. At the same time, for predictor set 2 (which incorporates, among others, information from the IMERG gridded precipitation dataset; see Table 1), gbm exhibits very similar performance to that of XGBoost. Additionally, the mean rankings (see Figure 6b) of random forests and XGBoost are of similar magnitude. In terms of the same criterion, gbm scores much closer to random forests and XGBoost than to linear regression.

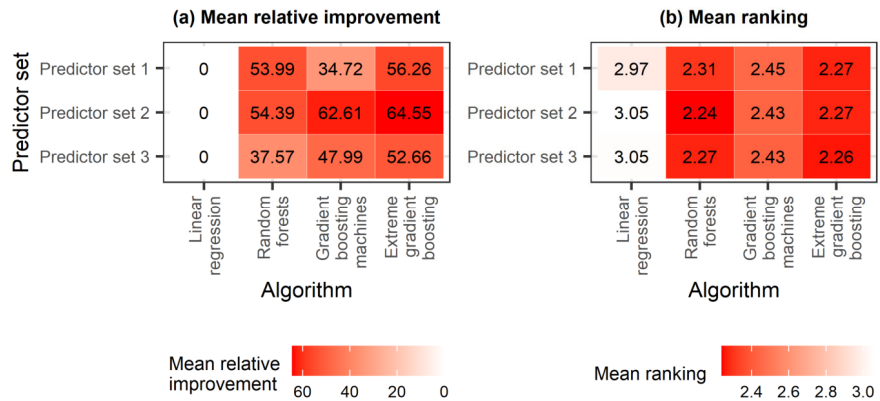

Figure 6. Heatmaps of the: (a) relative improvement (%) in terms of the median square error metric, averaged across the two folds, as this improvement was provided by each tree-based ensemble algorithm with respect to the linear regression algorithm; and (b) mean ranking of each machine and statistical learning algorithm, averaged across the two folds. The computations were made separately for each predictor set. The more reddish the colour, the better the predictions on average.

Moreover, Figure 7 facilitates more detailed comparisons with respect to the frequency with which each algorithm appeared in the various positions from the first to the fourth in the experiments. Here again, the comparisons can be made across both the algorithms and the predictor sets. The respective results are somewhat similar across predictor sets. Indeed, linear regression is much more likely to be found at the fourth (i.e., the last) place than at any other place. It is also more likely to be found at the first place than at the second and third places. At the same time, random forests are more likely to be ranked first than second, third and fourth, and gbm appears most often in the second and third positions. The last position is the least likely for both gbm and XGBoost. The latter algorithm is more likely to be ranked second; yet, the first and third positions are also far more likely than the last.

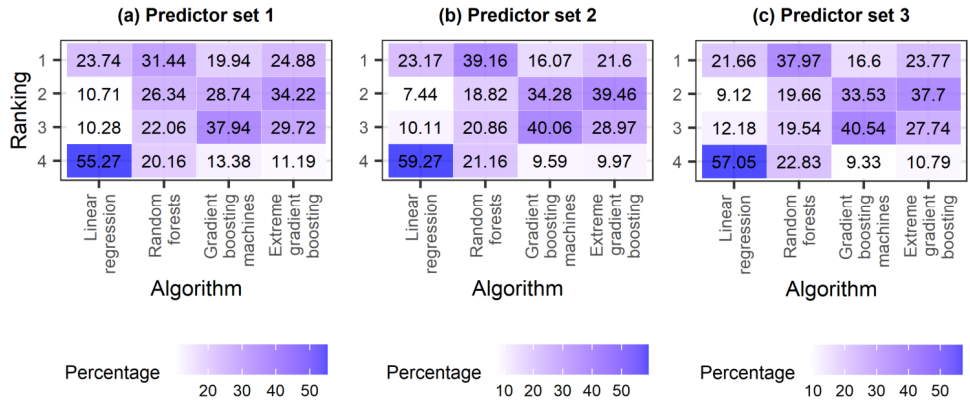

Figure 7. Heatmaps of the percentages (%) with which the four machine and statistical learning algorithms were ranked from 1 to 4 for the predictor sets (a–c) 1–3. The rankings summarized in this figure were computed separately for each pair {case, predictor set}. The darker the colour, the higher the percentage.

Lastly, Figures 8 and 9 allow us to compare the degree of information that is offered by the two gridded precipitation products within the context of our regression problem, further than the comparisons already allowed by the variable importance explorations using the gain metric incorporated into the XGBoost algorithm (see again Figure 5). Overall, the IMERG dataset was proven to be far more information-rich than the PERSIANN dataset, in terms of both mean relative improvement (see Figure 8a) and mean ranking (see Figure 8b). Indeed, the relative improvements with respect to the linear regression algorithm run with the predictor set 1 are much larger for the tree-based algorithms when these algorithms are run with predictor set 2 than when they are run with predictor set 1. Additionally, predictor set 3 (which contains information from both gridded precipitation datasets) does not improve the performances notably in terms of mean relative improvements with respect to predictor set 2, although it does in terms of mean ranking. While the best modelling choice is {XGBoost, predictor set 3}, random forests were ranked in the two first positions more often than any other algorithm for predictor sets 2 and 3, when the ranking was made collectively for all the predictor sets (see Figure 9). Still, for the same predictor sets, XGBoost appeared in the last few positions much less often and achieved the best performance in terms of mean ranking when run with predictor set 3.

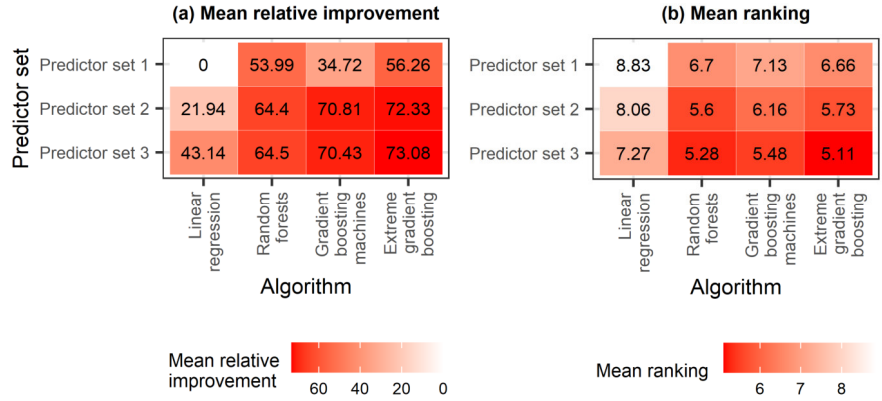

Figure 8. Heatmaps of the: (a) relative improvement (%) in terms of the median square error metric, averaged across the two folds, as this improvement was provided by each tree-based ensemble algorithm with respect to the linear regression algorithm, with this latter algorithm being run with the predictor set 1; and (b) mean ranking of each machine and statistical learning algorithm, averaged across the two folds. The computations were made collectively for all the predictor sets. The more reddish the colour, the better the predictions on average.

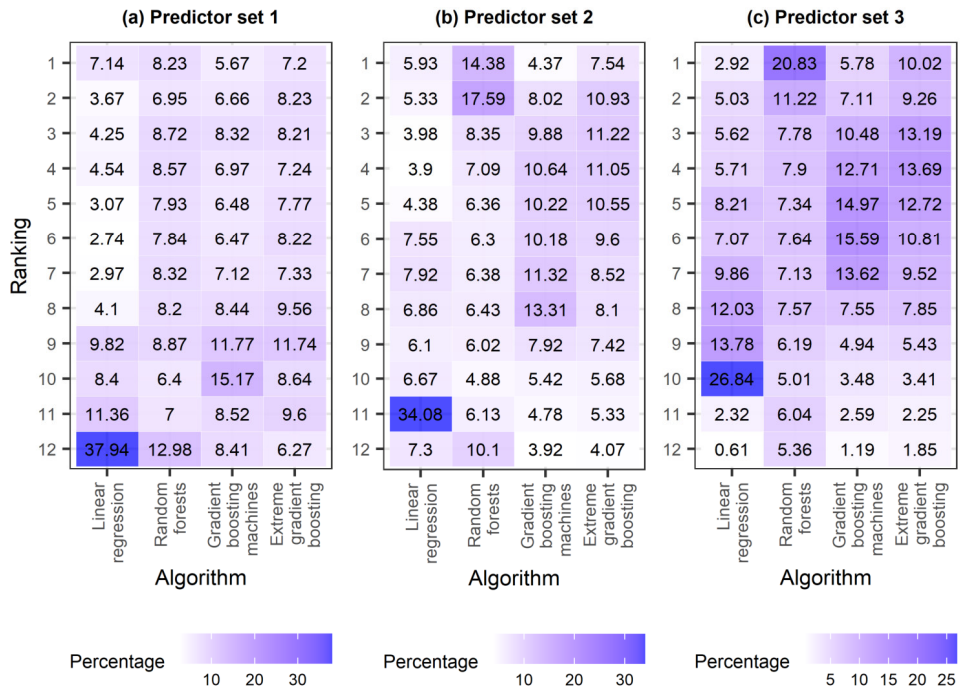

Figure 9. Heatmaps of the percentages (%) with which the four machine and statistical learning algorithms were ranked from 1 to 12 for the predictor sets (a–c) 1–3. The rankings summarized in this figure were computed separately for each case and collectively for all the predictor sets.

## 5. Discussion

Overall, XGBoost was proven to perform notably better than random forests and gbm when merging gridded satellite precipitation products and ground-based precipitation measurements for the contiguous US at the daily time scale. Notably, this result agrees with results offered by multiple competitions, in which large datasets from other applied disciplines were utilized, and where XGBoost outperformed other boosting algorithms and random forests. The better performance of XGBoost compared with gbm could be attributed to the fact that the former algorithm was armoured during its making with extra parameters compared with traditional boosting implementations, as well as with a regularization procedure for avoiding overfitting. Moreover, the variable importance scores obtained in this work and the same work's predictive performance comparison across predictor sets indicate that the IMERG product offers more useful predictors than the PERSIANN product for the same time scale. In summary, when the former of these products is utilized (either alone or together with the latter of them), random forests are far behind of both XGBoost and gbm in terms of accuracy. On the other hand, when PERSIANN is utilized, without IMERG being utilized as well, gbm is far behind of both XGBoost and random forests.

This latter result agrees, to some extent, with results obtained for the monthly time scale in Papacharalampous et al. (2023), although the relative scores with respect to the linear regression algorithm were found to be somewhat lower therein. Note, however, that the comparison in this latter work relied on the PERSIANN satellite dataset only. Still, it is accurate to deduce that the improvements in performance with respect to the linear regression algorithm offered by XGBoost, gbm and random forests, when all these four algorithms are run with the same predictor variables, are very large (i.e., from approximately 25% to approximately 65%) for both the daily and monthly time scales. Notably, even larger improvements could be achieved by combining predictions of diverge algorithms in advanced or even simple ensemble learning frameworks, following research efforts made in various fields (e.g., those by Bogner et al. 2017, Sagi and Rokach 2018, Papacharalampous et al. 2019, Tyralis et al. 2019a, Kim et al. 2021, Lee and Ahn 2021, Tyralis et al. 2021, Granata et al. 2022, Li and Yang 2022).

As the main concepts behind the boosting and random forest families of algorithms are different (see Section 2 for brief summaries of these concepts), their combinations could be investigated in the direction of achieving these further improvements with a low computational cost. Moreover, their combination with the linear regression algorithm could also be investigated. Indeed, in some contexts, even the least accurate algorithms could benefit ensemble learning solutions (see, e.g., the relevant comparison outcome in Papacharalampous and Tyralis 2020). In cases where the computational cost does not constitute a limiting factor in algorithm selection, neural network (Cheng and Titterington 1994, Jain et al. 1996, Paliwal and Kumar 2009) and deep learning (LeCun et al. 2015, Schmidhuber 2015) regression algorithms could be added to the ensembles. Lastly, instead of aiming at providing accurate mean-value predictions, one could aim at providing accurate median-value predictions coupled with useful uncertainty estimates. This would require working on machine and statistical learning methods, such as those summarized and popularized in the reviews by Papacharalampous and Tyralis (2022a) and Tyralis and Papacharalampous (2022a).

## 6. Conclusions

Precipitation datasets that simultaneously cover large regions with high density and are more accurate than satellite precipitation products can be obtained by correcting such products using earth-observed datasets together with machine and statistical learning regression algorithms. Tree-based ensemble algorithms are adopted in various fields for solving algorithmic problems with high accuracy and lower computational cost compared with other algorithms. Still, information on which tree-based ensemble algorithm to select when the merging is conducted for the contiguous United States (US) and at the daily time scale, at which the computational requirements might constitute a crucial factor to consider along with accuracy, is missing from the literature of satellite precipitation product correction.

Herein, we worked towards filling this methodological gap. We conducted an extensive comparison between three tree-based ensemble algorithms, specifically random forests, gradient boosting machines (gbm) and extreme gradient boosting (XGBoost). We exploited daily information from the PERSIANN (Precipitation Estimation from Remotely Sensed Information using Artificial Neural Networks) and the IMERG (Integrated Multi-satellitE Retrievals for GPM) gridded datasets, and daily earth-observed information from the Global Historical Climatology Network daily (GHCNd) database. The entire contiguous US was examined and results that are generalizable were obtained. These results indicate that XGBoost is more accurate than random forests and gbm. They also indicate that IMERG is more useful than PERSIANN in the context investigated.

**Conflicts of interest:** The authors declare no conflict of interest.

**Author contributions:** GP and HT conceptualized and designed the work with input from AD and ND. GP and HT performed the analyses and visualizations, and wrote the first draft, which was commented on and enriched with new text, interpretations and discussions by AD and ND.

**Funding:** This work was conducted in the context of the research project BETTER RAIN (BEnefiTTing from machine lEarning algoRithms and concepts for correcting satellite RAINfall products). This research project was supported by the Hellenic Foundation for Research and Innovation (H.F.R.I.) under the "3rd Call for H.F.R.I. Research Projects to support Post-Doctoral Researchers" (Project Number: 7368).

**Acknowledgments:** The authors are sincerely grateful to the Journal for inviting the submission of this paper, and to the Editor and Reviewers for their constructive remarks.

## Appendix A    Statistical software

We used the `R` programming language (R Core Team 2022) to implement the algorithms, and to report and visualize the results.

For data processing and visualizations, we used the contributed `R` packages `caret` (Kuhn 2022), `data.table` (Dowle and Srinivasan 2022), `elevatr` (Hollister 2022), `ncdf4` (Pierce 2021), `rgdal` (Bivand et al. 2022), `sf` (Pebesma 2018, 2022), `spdep` (Bivand 2022, Bivand and Wong 2018, Bivand et al. 2013), `tidyverse` (Wickham et al. 2019, Wickham 2022).

The algorithms were implemented by using the contributed `R` packages `gbm` (Greenwell et al. 2022), `ranger` (Wright 2022, Wright and Ziegler 2017), `xgboost` (Chen et al. 2022c).

The performance metrics were computed by implementing the contributed `R` package `scoringfunctions` (Tyralis and Papacharalampous 2022a, 2022b).

Reports were produced by using the contributed `R` packages `devtools` (Wickham et al. 2022), `knitr` (Xie 2014, 2015, 2022), `rmarkdown` (Allaire et al. 2022, Xie et al. 2018, 2020).